\documentclass[anon,12pt]{format} % Anonymized submission
\usepackage{float}
\usepackage{textgreek}
\usepackage{amsmath}
\usepackage{amssymb}
\usepackage{algorithm}
\usepackage{algorithmic}
\usepackage{mathtools}

\usepackage[english]{babel} 						
\usepackage{acronym}                    % Acronyms
\usepackage{color}                      % Enables defining of colors via \definecolor
\usepackage{listings}                   % Nicer source code listings
\usepackage{multicol}										% Content of a table over several columns
\usepackage{multirow}										% Content of a table over several rows
\usepackage{rotating}										% Alows to rotate text and objects

\title[The DSTC with Bayesian approach]{The Dialog State Tracking Challenge with Bayesian Approach}
\usepackage{times}
 \coltauthor{
 \Name{Quan Nguyen} \Email{5qnguyen@informatik.uni-hamburg.com}\\
 \addr Department of Informatics, University of Hamburg
 }

\begin{document}

\maketitle

\begin{abstract}
Generative model has been one of the most common approaches for solving the Dialog State Tracking Problem with the capabilities to model the dialog hypotheses in an explicit manner. The most important task in such Bayesian networks models is constructing the most reliable user models by learning and reflecting the training data into the probability distribution of user actions conditional on networks' states. This paper provides an overall picture of the learning process in a Bayesian framework with an emphasize on the state-of-the-art theoretical analyses of the Expectation Maximization learning algorithm. 
\end{abstract}


\newpage
\section{Introduction}

The problems of understanding users' intention has long been pursued by engineers and scientists in speech processing~\cite{Thomson2010EP}. Why is that such a hard problem? One main reason is that for a long time there are no effective dialog models that could match speech signals to some proper hypotheses about what the speakers are intending to do.

\begin{figure}[htb]
	\centering
    \includegraphics[width=.9\linewidth]{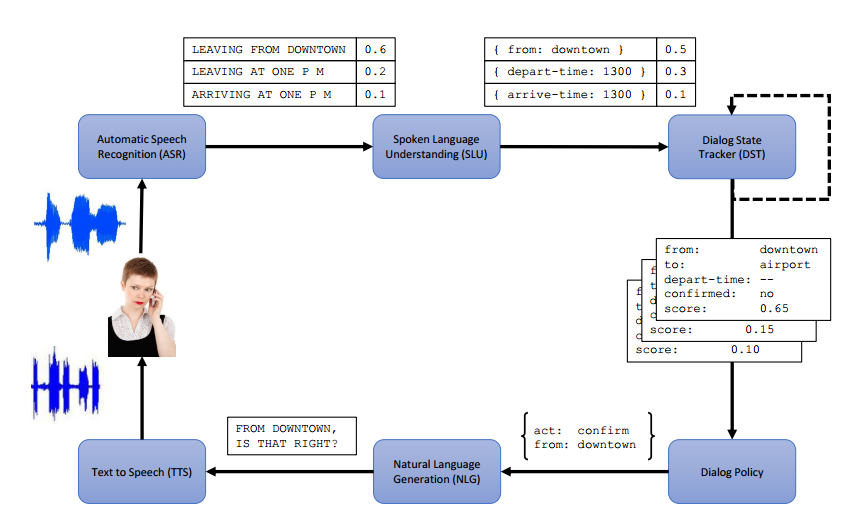}
    \caption{End-to-End model of a dialog system~\cite{williams2016dialog}}
    \label{img:1}
\end{figure}

Figure~\ref{img:1} illustrates one effective model suggested by~\cite{williams2016dialog}. Basically any dialog model needs to be capable of handling the following three tasks:

\begin{itemize}
\item Understanding the meaning of users' utterance given the current state of the dialog.
\item Understanding the changes of dialog's states given the meaning of users' utterance.
\item Taking appropriate actions based on the new states of the dialog.
\end{itemize}

It can be observed that the model in figure~\ref{img:1} has dedicated modules to fulfill all those requirements. Since the analysis results of one task is the input for the subsequent tasks, the accuracy of the first two tasks is crucial for a dialog system to derive suitable actions every time the system takes initiative.
 
Assume that we have already had a functioning  system and we want to find the bottleneck of the system. In other words, we want to know what would be the source of errors. Consider how the speech signal is transformed throughout the system: firstly the Automatic Speech Recognition (ASR) transcribes the sound into words, then the Speech Language Understanding unit (SLU) constructs the perceived words into small semantic units. The Dialog Manager (DM) takes in these semantic units and tries to fill in a number of slots. If the list of words the ASR churns out are highly incorrect, the subsequent modules (SLU and DM) are often mislead into erroneous understanding and actions.

Unfortunately, high error rate of ASR module is still common~\cite{williams2016dialog}. Consequently, in order to achieve a certain level of robustness, the SLU and especially DM urge to find measures to overcome the non-optimal output of ASR.

To address the above issues of ambiguity and misunderstanding, one main principle is implemented in the dynamics of the above three modules: maintaining multiple hypotheses on a probabilistic manners. Each hypothesis has an associate weight value that indicates the level of certainty that the system has on that hypothesis. For example, in the Figure~\ref{img:1} above the ASR thinks that it is most likely that the speaker said something about leaving downtown with a probability of 60 percent. Similarly, multiple instances of flight reservations (the most important semantic information) are stored in the DM in deciding which Dialog Policy should follow in the next step.

% Acknowledgments---Will not appear in anonymized version
% \acks{We thank a bunch of people.}

\section{The Dialog State Tracking Challenge}

The Dialog State Tracking Challenge (DSTC) provides a common testing framework for dialog state trackers. The main idea behind this contest-format testing framework is that for the same training and testing data, various trackers built upon different models compete together to find out the best model (i.e. highest performance score) in different scenarios and testing schemes (i.e. mis-match distribution between train and test data, changes of user's goals, open versus closed dictionary, etc)~\cite{williams2016dialog}.

\begin{figure}[htb]
	\centering
    \includegraphics[width=.9\linewidth]{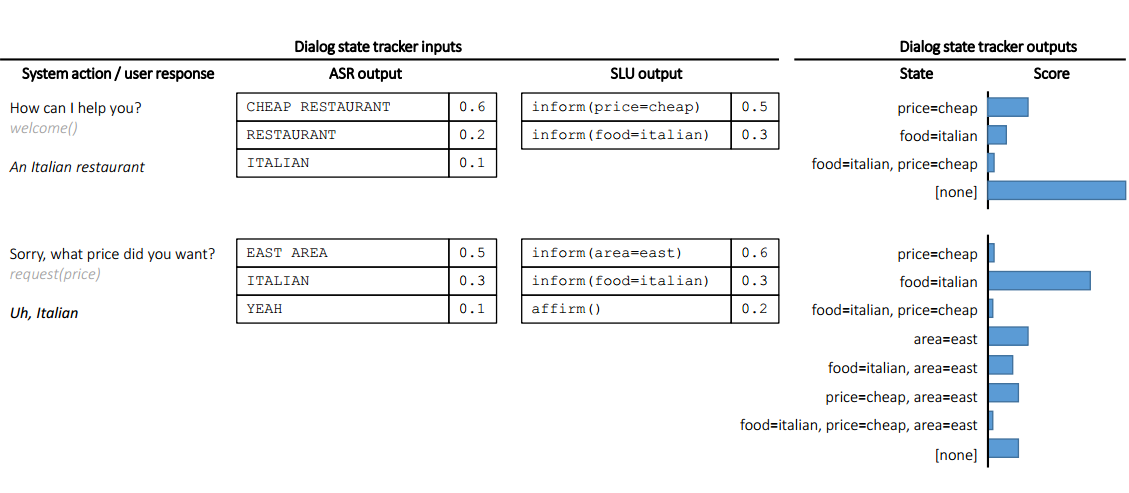}
    \caption{Based on the output of ASR and SLU modules, the dialog state tracker establish and maintain multiple hypotheses of the user's intention. The probability distribution of these hypotheses is subsequently used for deciding agent's actions~\cite{williams2016dialog}}
    \label{img:2}
\end{figure}

Figure~\ref{img:2} demonstrates the typical output of a tracker's output. A number of different (both contradicting and complementary) states are maintained and scored in the tracker. These scores are effectively constructed from not just a single speaking phase but multiple phases. As the dialog progresses, it is an expected behavior that the DM gradually update the scores of the states and the most probable states become more and more obvious. In this setting although the output of ASR is not very accurate, the system is capable of combining multiple outputs and selectively eliminating the most unlikely states. This behavior is very similar to that of Particle Filter where the current position of the robot become more and more apparent as the robot gets more information about the environment.

As mentioned previously, the system mainly operations in probabilistic manner with some stochastic models corresponds to how dynamics of a module leads to the states of the subsequent modules. There are two popular models emerge in this setting: the Discriminative model and Generative model. This paper only presents the learning algorithm in the Generative model employ by~\cite{Williams:2007:POM:1221595.1221967}.

\section{The Bayesian Method}

In generative models, the probability inference of the observation is formulated as a stochastic sequence generation mechanism from some latent variables. One simple yet effective model is Hidden Markov Model where the observation and transition states are ruled by Markov property in which, the probability of encountering the current state or observation depends only on the immediate previous state.

\begin{figure}[htb]
	\centering
    \includegraphics[width=.75\linewidth]{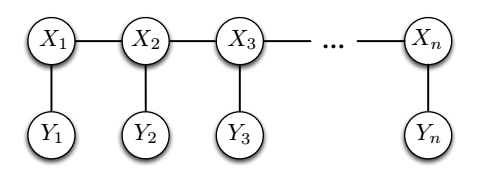}
    \caption{A Markov chain with $X$ being the hidden states and $Y$ being the observable states. The Markov property indicates that the likelihood of all observed states at time $t$ can be attributed to the single hidden state of time $t-1$. ~\cite{MIT2014ForwardBackward}}
    \label{img:3}
\end{figure}

Figure~\ref{img:3} illustrates the graphical model of a simple Hidden Markov Model (HMM). The sequence of hidden states and observations is called Markov chain. There are two main objectives of such model: the first one is to find the most likely hidden states that results in a given sequence of observations, the second one is to construct the best transition and observation probability from a randomly initialized setting. The accuracy of the former objective is entirely dependent on the accuracy of the latter objective. Therefore many system put a strong emphasize on evaluating and estimating these probabilities in the HMM.

In the context of spoken dialog system, the speech signal is the "observation" and the underlying hypotheses and states of dialog are the "latent variables". Since the the sequence of events is generated based on hidden variables, the system wants to explicitly formulate the probability that it will "hear" something given the current understanding of the dialog.

In the next section~\ref{section:EM}, we will discuss in detail the Expectation Maximization algorithm as the main learning methods in a generative model for DSTC.

\section{Expectation Maximization Algorithm}
\label{section:EM}
The Expectation Maximization algorithm (EM) is one the most commonly used optimization algorithm~\cite{Syed2008EMMain}. The main idea of EM is to find some appropriate probability distribution of the latent variables and based on that distribution, the parameters are repeatedly estimated with a better values than the previous ones. The objective function of EM is the log-likelihood function of the observed data. 

\subsection{Jensen Inequality}
Generally, Jensen Inequality states that if a function is convex, then the function of the expectation is always smaller than or equal to the expectation of that function~\cite{Borman2004expectation}.

\begin{equation}
	f(E[x]) \leq E[f(x)]
\end{equation}

The same property can be stated for a concave function with reversed inequality (greater than or equal instead of smaller than or equal). This strict evaluation (smaller or greater) of two terms: expected value of a function and that function's value at the expected value of its domain can be proved based on the convexity of the function~\cite{Syed2008EMMain}

\begin{figure}[htb]
	\centering
    \includegraphics[width=.8\linewidth]{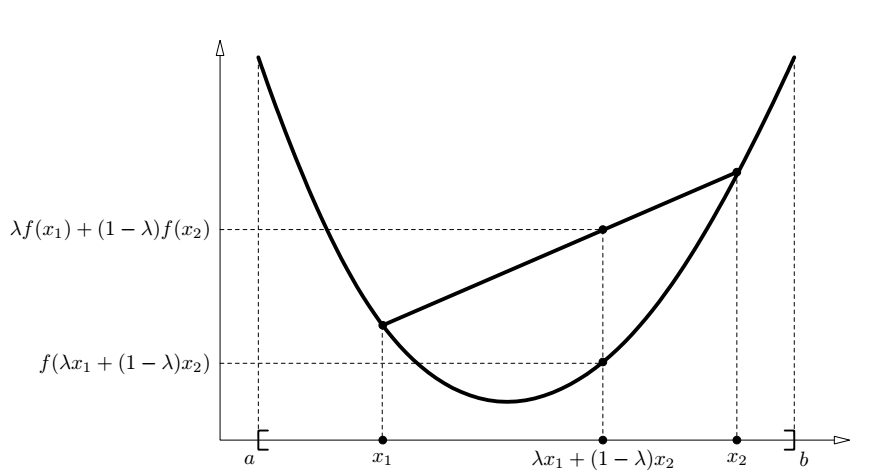}
    \caption{An example of convex function $f$. Function $f$ is guaranteed to be convex if the expected value of its function is larger than its value on the expected value of its domain.~\cite{Borman2004expectation}}
    \label{img:4}
\end{figure}

Figure~\ref{img:4} illustrates the relative comparison between the two quantities in a parabolic convex function. It is observable (and mathematically provable) that the equality in Jensen inequality holds only for the case when the variable is identical with its expected value

\begin{equation*}
X = E[x]
\end{equation*}

Why is this inequality useful? Consider the logarithm function

\begin{equation*}
f(x) = ln(x)
\end{equation*}

Its second-order derivative is

\begin{equation*}
f''(x) = -\frac{1}{x^2} < 0
\end{equation*}

This indicates that the logarithm function is concave and so is the objective log-likelihood function. This fact enables us to apply Jensen inequality on the objective function and attain a tractable \textbf{lower bound} of the objective function.

\subsection{Expectation Maximization Algorithm}

The reason behind finding a lower bound estimation of the objective function lies in the intractability of the objective function itself. Since the distribution of latent variables is unknown in most of the case, directly maximizing the objective function by traversing through all possible configuration of hidden states is downright unfeasible. A better method would be finding a strict lower bound function and maximizing the lower bound function instead~\cite{Borman2004expectation}. The most crucial property that this method need to possess is the convergence of the final state. In other words, it is absolutely required that the after each optimization step, the new parameters is strictly better than the previous one.

\begin{figure}[htb]
	\centering
    \includegraphics[width=.8\linewidth]{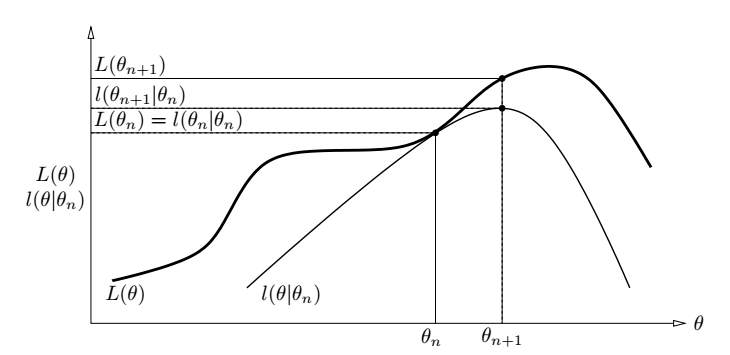}
    \caption{Optimizing an (intractable) objective function by optimizing its lower bound. The next parameter at time step $t+1$ has to be chosen selectively to ensure the convergence of the algorithm~\cite{Borman2004expectation}}
    \label{img:5}
\end{figure}

Figure~\ref{img:5} demonstrates one optimization step in EM algorithm. Notice when the algorithm moves from the current parameter $\theta_n$ to the new one $\theta_{n+1}$, the actual likelihood function $L$ increases following the increasing of the lower bound function $l(\theta|\theta_n)$. After the lower bound function has achieved the maximal values, EM algorithm will stop although the actual likelihood still increases. This stop condition ensure that function $L$ is monotonic and guaranteed a (local maximal) convergence.

\begin{algorithm}
  \caption{Find MLE by EM}
  \begin{algorithmic}
  \REQUIRE Domain-space is well-defined
  \REQUIRE Random initialization procedure of parameters vector is available
  \STATE $T = $ maximum iterations
  \STATE $\theta = $ randomly initialized
  \FOR {t=1,...,T}
 \STATE $\theta^{t} = argmaxE_y[logPr(X, Y | \theta) | Y, \theta^{t-1}]$
 \ENDFOR
 \RETURN $\theta^T$ and $Pr(Y | \theta^T)$
  \end{algorithmic}
\end{algorithm}

\textbf{Algorithm 1} is a step-by-step illustration of EM algorithm. Formally, the algorithm incrementally looks for the most optimal configuration of parameters. The use of Jensen Inequality allows us to transform the initial intractable optimization problem to a new tractable problem at the cost of losing the generality in finding the global maximum. Nevertheless it has been proved empirically that the EM algorithm achieves very good performance in the DSTC~\cite{williams2016dialog}.

\subsection{Forward-Backward Algorithm}

In the process of finding the optimal parameters in EM algorithm, one frequent sub-procedure is to calculate the probability of obtaining a subset of events or the whole events given some values of the states. This computation is a non-trivial task since it requires some manipulation over the graphical model depicted by the HMM above. This section provide an overview on how to perform such computation not only on the HMM itself but on general acyclic graphs. The main principle behind this computation is Dynamic Programming which is an tractable method to calculate any values of states given all the causal transition states before~\cite{MIT2014ForwardBackward}.

\begin{algorithm}
  \caption{Forward-Backward algorithm}
  \begin{algorithmic}
  \REQUIRE Probability distribution of initial unobserved states is well-defined
   \STATE $Y = $ set of all $N$ observed events
  \STATE $X = $ set of all $N$ latent variables
  \STATE Run Forward algorithm
  \STATE Run Backward algorithm
  \FOR{k = 1,...,N}
  \STATE compute all $Pr(X_k | Y)$
  \ENDFOR
  \end{algorithmic}
\end{algorithm}

\textbf{Algorithm 2} is a demonstration of how the Forward-Backward algorithm is implemented in general. As its name suggested, the algorithm requires three runs over all states, with the last run combines the results from the previous two. The aim of the final run is to calculate the likelihood probability of having a sequence of hidden states from the beginning to each hidden state. To obtain that result, the algorithm needs two information: the joint probability of these two terms and the posterior probability of the observed events given a prefix of sequence of hidden states.

\begin{figure}[htb]
	\centering
    \includegraphics[width=.75\linewidth]{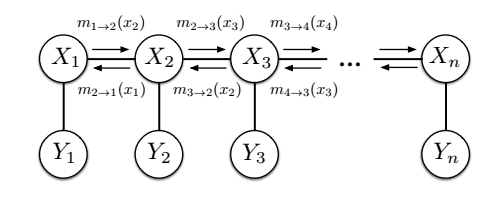}
    \caption{Markov chain with Forward and Backward flow of probabilities. \cite{MIT2014ForwardBackward}}
    \label{img:6}
\end{figure}

An illustration of the Forward-Backward algorithm can be found in Figure~\ref{img:6}. Since the graph has no cycle, any arbitrary node on the graph contains all the information of its ancestors (depends on which direction the algorithm is running, the ancestors could be the previous or subsequent hidden states and observations). The algorithm basically consider one node at one time and never go in reverse direction. This key observation is crucial in making the algorithm tractable.

\begin{algorithm}
  \caption{Forward: compute joint probability of both observed and unobserved states}
  \begin{algorithmic}
  \REQUIRE Transition probabilities are well-defined
  \REQUIRE Prior distribution is well-defined
  \REQUIRE Probability distribution of initial unobserved states is well-defined
  \STATE $Y = $ set of all $N$ observed events
  \STATE $X = $ set of all $N$ unobserved events
  \FOR{k = 1,...,N}
  \STATE Recursively compute $Pr(X_k, Y)$ by Dynamic Programming algorithm
  \ENDFOR
  \end{algorithmic}
\end{algorithm}

\begin{algorithm}
  \caption{Backward: compute likelihood of a range of observed states given a single prior unobserved state}
  \begin{algorithmic}
   \REQUIRE Transition probabilities are well-defined
  \REQUIRE Prior distribution is well-defined
  \REQUIRE Probability distribution of initial unobserved states is well-defined
  \STATE $Y = $ set of all $N$ observed events
  \STATE $X = $ set of all $N$ unobserved events
  \FOR{k = 1,...,N}
  \STATE Recursively compute $Pr(Y_{k+1 : N}| X_k)$ by Dynamic Programming algorithm
  \ENDFOR
  \end{algorithmic}
\end{algorithm}

\textbf{Algorithm 3} and \textbf{Algorithm 4} illustrates the simplicity in implementation of the two procedure: Forward run and Backward run. Indeed, it is the simplicity and efficiency of the algorithm being one of the reason that make it popular in every circumstances when the probabilistic graphical model is a acyclic graph. 

\section{Empirical results}
\label{section:ER}
\begin{figure}[htb]
	\centering
    \includegraphics[width=.75\linewidth]{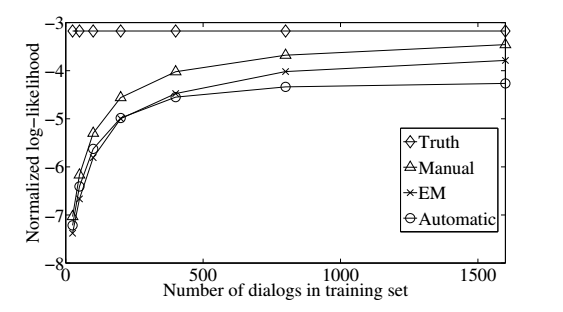}
    \caption{Performance of EM is improved with increasing size of training set~\cite{Syed2008EMMain}}
    \label{img:7}
\end{figure}

\begin{figure}[htb]
	\centering
    \includegraphics[width=.75\linewidth]{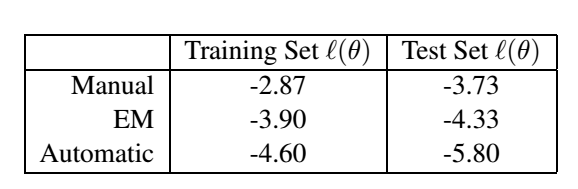}
    \caption{The normalized log-likelihood of three different algorithms. EM shows an improvement over automatic transcribed dialog but still behind manually transcribed dialog~\cite{Syed2008EMMain}.}
    \label{img:8}
\end{figure}

The empirical performance of EM algorithm in comparison with the two other transcribed dialog methods can be found in Figure~\ref{img:7} and Figure~\ref{img:8}.
In general, it can be observed that EM works better than Automatic transcribed logs but worse than Manual transcribed logs.

The learning curve depicted in Figure~\ref{img:7} indicates a monotonic increasing relationship between performance of an algorithm and number of dialog in training set. The justification is obvious: with more data in training set, the closer the estimated model to the optimal setting. The exact log-likelihood value of each method can be found in Figure~\ref{img:8}.

The discrepancy between manual and automatic transcribed logs can be explained by the erroneous of ASR module. Since ASR is not optimal, a system without Dialog Manager will perform worse than a system with optimization step like Bayesian method. For the same reason, the manual transcribed method apparently eliminate all possible errors from the ASR and thus achieve the best result among the three. The aim of research in the field is to get the performance of generative model closer and closer to the manual transcribed method.

\section{Conclusion and Discussion}

The convergence of EM algorithm has been proved in~\cite{Collins1997}. Another proof can be found in~\cite{EMDemystified}. However, the gradual optimization in EM is only as good as gradient descent which makes it prone to saddle points~\cite{Collins1997}. It should be noted by gradient descent, we are referring to the optimization performed on the original likelihood function by calculating the derivative of log-likelihood function and add the derivatives to the parameters, similar to the back-propagation learning algorithm in neural networks.

The inherent weakness of generative model is the necessary to model the prior distribution of latent variables $p(\theta)$. While in some circumstances modeling this prior distribution could be beneficial in the sense that it tells us how the latent variables are spanned in their domain space, we can hardly have enough data and computational resources to accurately estimate this distribution. Indeed it has been proven in~\cite{williams2016dialog} that in all three DSTC the discriminative models always outperform the generative models by a large margin. However, it should be noted that the superior of discriminative models come in the condition of enough volume of data. In the cases where data is not enough to build a good model, discriminative models are easily overfitting while unable to tells us any meaningful information about the nature of the system.

As shown in section~\ref{section:ER}, the performance of generative models are far from the manual transcribed dialog and the absolute truth. While a better ASR will certainly increases the performance of the whole system, building a better model to exploit the output of ASR and SLU is still an active research field. We have seen above that the performance of the model increases by training on more and more data, so incorporating the system into a big data architecture with proper scaling could be one promising measure in the way to achieve a human-like performance of dialog models. Another method which includes rigorous mathematical analysis is to find tighter lower bound estimations for the likelihood. While the Jensen inequality has proven to be able to achieve reasonable results, having a stricter evaluation on the lower bound will certainly benefits the optimization process by increasing the optimal values of converged states and allowing longer training time for better use of the increasing amount of data and computational powers.

\newpage
\bibliography{bib}

\end{document}